\documentclass[conference]{IEEEtran}
\IEEEoverridecommandlockouts
\usepackage{cite}
\usepackage{amsmath,amssymb,amsfonts}
\usepackage{algorithmic}
\usepackage{graphicx}
\usepackage{array}
\usepackage{booktabs}
\usepackage{placeins}
\usepackage{float}
\usepackage{array}
\usepackage{siunitx}
\usepackage{booktabs}
\usepackage{textcomp}
\usepackage{caption} 
\usepackage{xcolor}
\def\BibTeX{{\rm B\kern-.05em{\sc i\kern-.025em b}\kern-.08em
    T\kern-.1667em\lower.7ex\hbox{E}\kern-.125emX}}
\begin{document}

\title{SFA\-Net: Spatial-Frequency Attention Network for Deepfake Detection}





\author{

\IEEEauthorblockN{Vrushank Ahire\textsuperscript{1} \hspace{2.5cm} Aniruddh Muley\textsuperscript{2} \hspace{2.5cm} Shivam Zample\textsuperscript{2}}
\IEEEauthorblockA{\textit{2022csb1002@iitrpr.ac.in} \hspace{1.5cm} \textit{2022mcb1257@iitrpr.ac.in} \hspace{1.5cm} \textit{2022mcb1280@iitrpr.ac.in}}

\vspace{0.5cm} 

\IEEEauthorblockN{Siddharth Verma\textsuperscript{1} \hspace{1.9cm} Pranav Menon\textsuperscript{1} \hspace{2.2cm} Surbhi Madan\textsuperscript{1,$\dagger$} \hspace{1.95cm}  Abhinav Dhall\textsuperscript{3,$\ddagger$}}
\IEEEauthorblockA{\textit{2022csb1126@iitrpr.ac.in} \hspace{0.65cm} \textit{2022csb1329@iitrpr.ac.in} \hspace{0.65cm} \textit{surbhi.19csz0011@iitrpr.ac.in} \hspace{0.65cm} \textit{abhinav.dhall@monash.edu}}

\vspace{0.5cm} 

\IEEEauthorblockA{
\textsuperscript{1}Department of Computer Science and Engineering, Indian Institute of Technology Ropar, Punjab, India \\
\textsuperscript{2}Department of Mathematics and Computing, Indian Institute of Technology Ropar, Punjab, India \\
\textsuperscript{3}Faculty of Information Technology, Monash University, Melbourne, Australia}
\IEEEauthorblockA{
\{\textsuperscript{$\dagger$}Mentor, \textsuperscript{$\ddagger$}Supervisor\}
}
}

\maketitle

\begin{abstract}
Detecting manipulated media has now become a pressing issue with the recent rise of deepfakes. Most existing approaches fail to generalize across diverse datasets and generation techniques. We thus propose a novel ensemble framework, combining the strengths of transformer-based architectures, such as Swin Transformers and ViTs, and texture-based methods, to achieve better detection accuracy and robustness. Our method introduces innovative data-splitting, sequential training, frequency splitting, patch-based attention, and face segmentation techniques to handle dataset imbalances, enhance high-impact regions (e.g., eyes and mouth), and improve generalization. Our model achieves state-of-the-art performance when tested on the DFWild-Cup dataset, a diverse subset of eight deepfake datasets. The ensemble benefits from the complementarity of these approaches, with transformers excelling in global feature extraction and texture-based methods providing interpretability. This work demonstrates that hybrid models can effectively address the evolving challenges of deepfake detection, offering a robust solution for real-world applications.
\end{abstract}


\section{Introduction}
The rapid advancement of deep learning and generative models has led to the proliferation of deepfakes. AI-generated images, videos, and audio recordings are becoming increasingly realistic, making it difficult for humans and traditional systems to distinguish between real and manipulated content. Deepfakes are often misused for acts such as identity theft\cite{b22} and spreading misinformation \cite{b23}, thus eroding trust in digital media. As a result, developing robust and reliable deepfake detection methods has gained high priority in cybersecurity research.

Researchers have proposed a variety of approaches to identify manipulated content over the years \cite{b24}. For example, MesoNet \cite{b1} focuses on mesoscopic features and low-level visual cues to identify fake images. Other models use texture-based features such as Local Binary Patterns (LBP) and the Fast Fourier Transform (FFT) \cite{b25} to detect inconsistencies in manipulated images. Transformers \cite{b16} constitute the state-of-the-art of modern deep learning architecture, with Vision Transformers (ViT) \cite{b20} and Swin Transformers \cite{b5} using global and local feature representations for face-related applications \cite{b26}.

Despite these advancements, deepfake detection remains a very challenging task due to the evolving nature of generative technology. Techniques such as Stable Diffusion \cite{b27} and Generative Adversarial Networks (GANs) \cite {b28} continue to produce increasingly realistic synthetic images. As a result, detection methods need to be continuously improved. Additionally, models fail to generalize successfully, as models trained on one method of generation often struggle to perform well on others \cite{b11}. Hence, state-of-the-art models require high accuracy in detecting forged content while having the ability to generalize well and perform on unseen data.

\begin{figure}[htbp]
    \centering
    \includegraphics[width=0.4\textwidth]{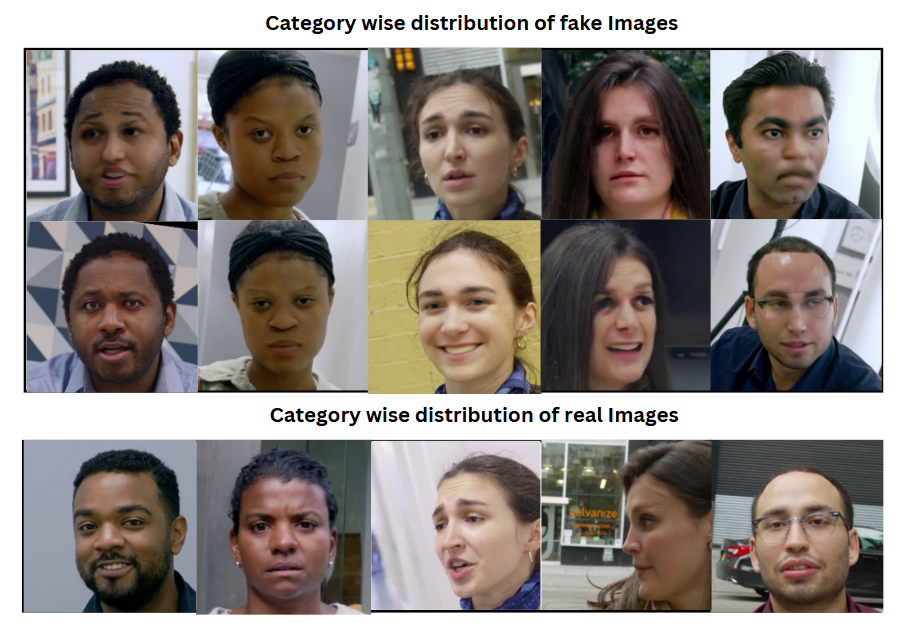} 
    \caption{Category wise segregation of images.}
    \label{fig:Fig1}
\end{figure}

To identify deepfakes, we experimented with various state-of-the-art CNN-based and Transformer-based architectures, both independently and in hybrid configurations. We evaluated different data-splitting techniques, as shown in Figure \ref{fig:Fig1}, to enhance generalization and improve model performance. Based on our findings, we propose an ensemble model that combines the strengths of the best-performing models and pipelines. These Transformer-based architectures compensate the shortcomings of each individual model, while providing high accuracy and robustness.


The rest of the paper is organized as follows: Section II provides a comprehensive review of related works, Section III details the proposed methodology, Section IV describes the experiments and results, Section V concludes the paper, highlighting key findings and suggesting future research directions followed by references used.

\section{Related Works}

The field of deepfake detection has seen significant advancements, with researchers proposing diverse methodologies to address the challenges of identifying manipulated content. Afchar et al. proposed MesoNet \cite{b1}, a compact neural network designed for facial video forgery detection. By focusing on mesoscopic features, MesoNet achieves 97.3\% accuracy on the FaceForensics++ dataset. This method is one of the earliest and simplest approaches, relying on low-level visual cues to detect manipulated content.

Following on these early efforts, Hu et al. introduced Squeeze-and-Excitation Networks (SENet) \cite{b2}, which incorporates a SE block which is a lightweight module that recalibrates channel-wise features. SENet is integrated into the architectures like ResNet. SENet reduces ImageNet top-5 error by 25\%, which is the state-of-the-art performance. This method advances beyond texture-based approaches by dynamically weighting feature channels to refine detection accuracy. Further expanding on the idea of feature enhancement, Zhang et al. proposed WS-DAN \cite{b3}, which is a fine-grained visual classification method that uses attention maps for guided data augmentation. By employing methods like attention cropping and attention dropping, WS-DAN can focus on discriminative parts of an object, which improves feature extraction and classification without requiring additional annotations. 

Building on top of these innovations, Zhuang Liu et al. introduced ConvNeXt \cite{b4}, which is a modernized ConvNet which is inspired by Transformer designs. By incorporating larger kernels and LayerNorm, ConvNeXt achieves 87.8\% top-1 accuracy on ImageNet and even outperforms Swin Transformers on the COCO detection and ADE20K segmentation. This hybrid approach bridges the gap between convolutional networks and transformers, offering significant performance gains for deepfake detection.

As transformer-based architectures gained prominence, Liu et al. proposed the Swin Transformer \cite{b5}, which is a hierarchical vision transformer that computes self-attention within shifted non-overlapping windows. This design achieves linear computational complexity, making it efficient for large-scale image analysis. The Swin Transformer represents a shift from convolutional networks to transformer-based architectures, enabling better handling of global and local features for deepfake detection. Building upon the transformer efficiency, DeiT \cite{b6} (Data-efficient Image Transformers) optimizes training using knowledge distillation and data augmentation, reducing dataset dependency. It can achieve up to 85.2\% top-1 accuracy on ImageNet-1k. DeiT further refines Swin Transformer’s advantages by improving training efficiency and minimizing reliance on large datasets, reinforcing its applicability in deepfake detection.

Targeting identity-specific anomalies, Huang et al. \cite{b7} introduced an identity-driven deepfake detection framework that identifies implicit inconsistencies in face-swapping manipulations. Achieving 98.5\% accuracy on the FaceForensics++ dataset, this method offers a specialized solution by focusing on identity-related artifacts, thereby enhancing generalization across different manipulation techniques.

Apart from this people have also tried to focus on surface-level analysis, SurFake \cite{b8} utilizes surface anomalies in images to detect deepfakes. By analyzing the Global Surface Descriptor (GSD), which captures geometric features of surfaces, this approach detects inconsistencies introduced by deepfake generation. This approach combines GSD with RGB data to improve detection, offering a novel perspective on deepfake identification by focusing on geometric inconsistencies.

Researchers have also tried to use physiological signals to their advantage. FakeCatcher \cite{b9} detects synthetic portrait videos by analyzing biological signals such as heart rate, which generative models struggle to replicate accurately. By assessing spatial coherence and temporal consistency, FakeCatcher provides a robust and model-agnostic detection approach, surpassing surface-based methods by using physiological cues that are inherently difficult to forge.

Addressing domain conflicts in multi-dataset training, GM-DF \cite{b10} proposed a Generalized Multi-Scenario Deepfake Detection framework. By employing hybrid expert modeling for domain-specific features, CLIP for common feature alignment, and meta-learning for generalization, GM-DF achieves state-of-the-art performance across diverse datasets. This approach synthesizes multiple techniques to create a robust and generalizable deepfake detection framework, representing a significant advancement in the field.

\section{Methodology}

The dataset contains a mix of 8 publicly available datasets: Celeb-DF-v1, Celeb-DF-v2, FaceForensics++, DeepfakeDetection, FaceShifter, UADFV, Deepfake Detection Challenge Preview, and Deepfake Detection Challenge \cite{b11}. The dataset contains face crops of size 256x256 of frames from videos constituting the 8 datasets. The training and validation sets are divided into 2 subcategories: real and fake. The training set is skewed, with there being 42690 real images and 219470 fake images, meaning a fake-to-real ratio of roughly 5.14:1. In addition to utilizing the full training dataset, we explored alternative data processing techniques to mitigate imbalance and enhance generalization, aiming to improve the model's performance across diverse datasets and generation methods.

\subsection{ Human Feature-Based Data Segmentation}
Latest deepfake techniques often struggle to accurately replicate subtle color variations, skin tones, and emotional nuances, particularly in critical regions such as the eyes and mouth, where inconsistencies are more likely to manifest.
To identify local features across different types of images, an attempt was made to divide the complete dataset. A model called DeepFace \cite{b12} was used to predict the features of the person in the image. The image was categorized based on the following features:

\begin{itemize}
    \item Race of person, \{white and others\}
    \item Emotion of the person, \{happy, negative, neural, scared\}
\end{itemize}

The original DeepFace model identified several other emotions as well, but similar emotions were grouped together to ensure fairness across all classes. Similarly, using multiple races (White, Black, Middle Eastern, etc) resulted in small clusters, causing issues in model learning due to lack of data. The dataset was first divided based on race, and then on emotion. This division resulted in 8 distinct classes, and images passed through the DeepFace model were classified into these categories.

\begin{table}[htbp]
\centering
\renewcommand{\arraystretch}{1.5}
\begin{tabular}{|l|c|c|}
\hline
\textbf{Category } & \textbf{Total Train} & \textbf{Total Validation} \\
\hline
other\_happy & 17,038 (14,280 + 2,749) & 79 (31 + 49) \\
\hline
other\_negative & 40,263 (33,890 + 6,373) & 653 (429 + 224) \\
\hline
other\_neutral & 54,757 (46,174 + 8,583) & 554 (373 + 181) \\
\hline
other\_scared & 18,117 (15,240 + 2,877) & 77 (25 + 52) \\
\hline
white\_happy & 17,759 (14,801 + 2,958) & 109 (22 + 87) \\
\hline
white\_negative & 44,719 (34,310 + 6,650) & 783 (294 + 489) \\
\hline
white\_neutral & 40,960 (34,310 + 6,650) & 594 (280 + 314) \\
\hline
white\_scared & 25,803 (21,122 + 4,681) & 190 (54 + 136) \\
\hline
\end{tabular}
\vspace{3pt}
\caption{Summary of Data by Category}
\label{tab:data_summary}
\end{table}

On each dataset division, we trained a separate ViT. The results of the ViT were as follows:

\begin{table}[htbp]
\centering
\renewcommand{\arraystretch}{1.5}
\begin{tabular}{|l|c|c|c|}
\hline
\textbf{Category (race\_emotion)} & \textbf{Epoch} & \textbf{Validation Loss} & \textbf{Accuracy} \\
\hline
other\_happy & 8 & 0.2661 & 0.9241 \\
\hline
other\_negative & 9 & 0.3821 & 0.9158 \\
\hline
other\_neutral & 3 & 0.3140 & 0.9025 \\
\hline
other\_scared & 6 & 1.4107 & 0.7792 \\
\hline
white\_happy & 2 & 0.3924 & 0.8257 \\
\hline
white\_negative & 9 & 0.5717 & 0.8748 \\
\hline
white\_neutral & 4 & 0.5619 & 0.8704 \\
\hline
white\_scared & 4 & 0.6604 & 0.8158 \\
\hline
\end{tabular}
\vspace{3pt}
\caption{Performance Metrics by Category}
\label{tab:performance_metrics}
\end{table}

The weighted accuracy is calculated as:
\[
\text{Weighted Accuracy} = \frac{\sum_{i=0}^{8} n_i a_i}{\sum_{i=0}^{8} n_i},
\]
where \( n_i \) is the number of images in the \( i^{\text{th}} \) category, and \( a_i \) is the accuracy of the \( i^{\text{th}} \) category. This technique achieved a weighted accuracy of 0.8812.

The division of classes proved beneficial for certain categories, such as \texttt{other\_happy}, \texttt{other\_negative}, and \texttt{other\_neutral}. However, the model exhibited comparatively lower performance for categories like \texttt{white\_happy}, \texttt{other\_scared}, and \texttt{white\_scared}. This discrepancy in performance may be attributed to the imbalance between the number of real images and deepfakes within these specific classes.

\begin{figure}[htbp]
    \centering
    \includegraphics[width=0.5\textwidth]{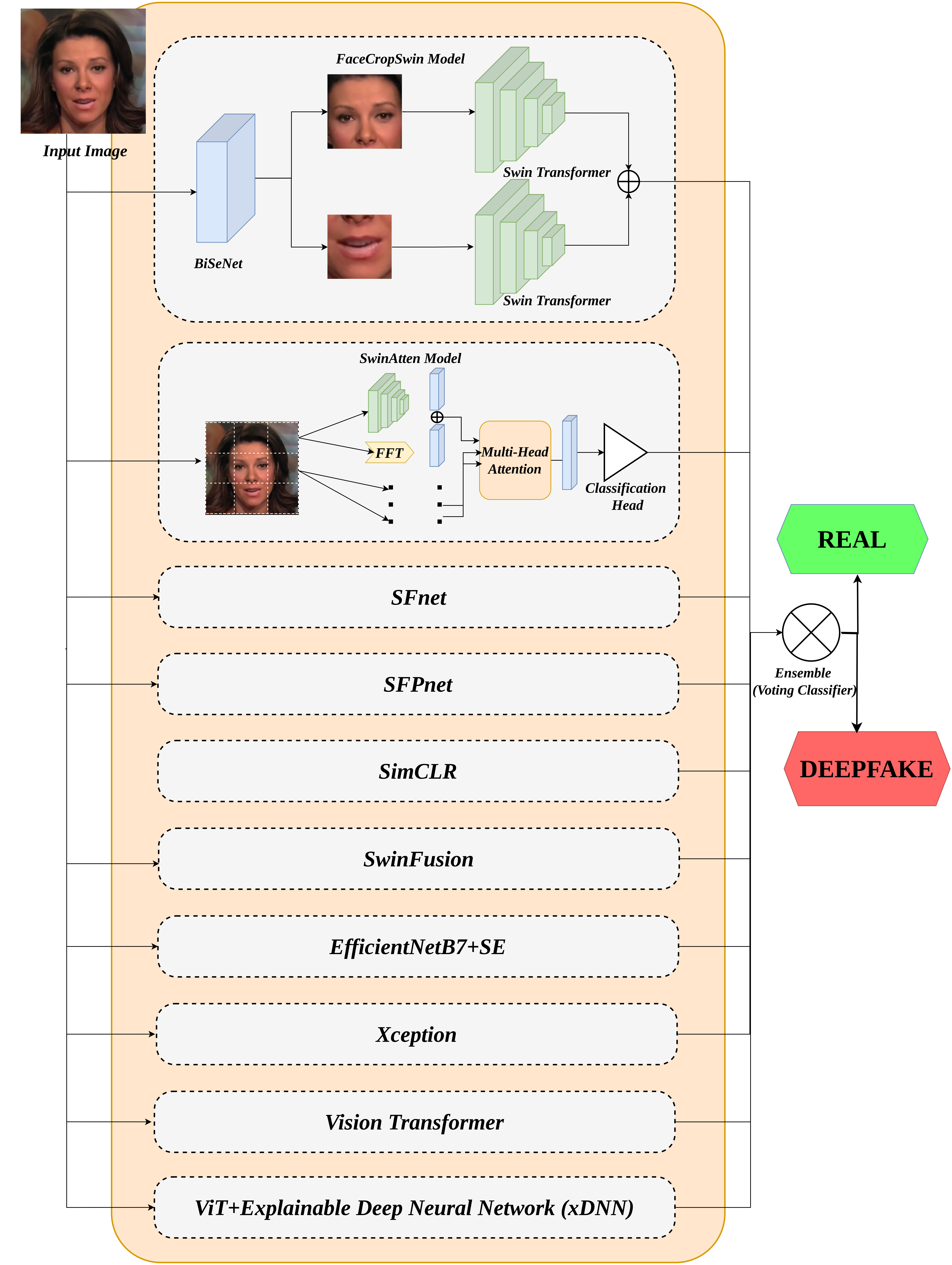} 
    \captionsetup{font=large} 
    \caption{Architecture of the proposed ensemble model for deepfake detection.}
    \label{fig:sample_image}
\end{figure}

\subsection{Fake Data Clustering}
To address data imbalance, we applied clustering to the fake training data. Given the fake-to-real dataset ratio of 5.14:1, we created 5 clusters using embeddings generated by the EfficientNet-B7 model \cite{b13} and applied k-means clustering. The real dataset remained undivided to maintain its integrity while balancing the distribution of fake data.

We initially employed multiple Swin Transformer models, with each model trained on the real dataset R paired with one of the five fake clusters (F1-F5). The embeddings from these models were concatenated and processed through an MLP classifier. However, this approach not only yielded unsatisfactory accuracy but also proved to be computationally expensive, leading us to discard it.

We then implemented a single-model solution with sequential training. The model was trained for three epochs on R and F1, followed by three epochs on R and F2, maintaining the weights between transitions. This process continued through all five fake clusters. The final phase involved fine-tuning on the complete dataset comprising all real and fake samples. The aim of this approach was to provide the model with a balanced learning environment in each ``fold" (the real set paired with one fake cluster), ensuring exposure to diverse data distributions. By fine-tuning on the full dataset afterward, the model retained a ``comprehensive view" of the data, mitigating the risk of catastrophic forgetting and enhancing its ability to generalize across the entire dataset.

We trained two models using this sequential training strategy: \textbf{SwinFusion}, based on the Swin Large architecture, and $XceptionFusion$, utilizing the Xception model \cite{b14}. 

\begin{table}[htbp]
\centering
\renewcommand{\arraystretch}{1.5}
\begin{tabular}{|l|c|c|}
\hline
\textbf{Model } & \textbf{Validation Loss} & \textbf{Validation Accuracy} \\
\hline
SwinFusion & 0.2830 & 0.9297 \\
\hline
XceptionFusion & 0.4439 & 0.9007 \\
\hline
\end{tabular}
\vspace{3pt}
\caption{Model Performances Using Data Clustering}
\label{tab:data_summary}
\end{table}

Accuracies exceeding 90\% indicate that our technique likely contributed to improved generalization, demonstrating the effectiveness of the sequential training and fine-tuning approach in enhancing model performance.

\subsection{Region-Based Facial Cropping}
Our third approach focused on the high-impact regions of deepfake images, particularly the eyes and mouth, as these areas are frequently targeted by deepfake manipulations and often contain detectable artifacts that aid in forgery detection. To achieve this, we utilized BiSeNet \cite{b15}, a pretrained face segmentation model, to segment the face into its constituent parts, including the skin, eyebrows (left and right), eyes, nose, lips (upper and lower), hair, ears, neck, and skin. If the image contained both lips, both eyes, and both eyebrows, we made 2 crops: one containing the eyes and eyebrows (and most often the full nose), and one containing both the lips (and by extension, the chin), and stored them. If any facial parts were missing due to blurring, manipulation, occlusion, or the subject facing away from the camera, the image was not cropped and was instead stored in a separate folder as a full image. This approach allowed us to concentrate on regions most likely to reveal deepfake artifacts while maintaining flexibility for cases where partial facial data was unavailable.

\begin{figure}[htbp]
    \centering
    \includegraphics[width=0.45\textwidth]{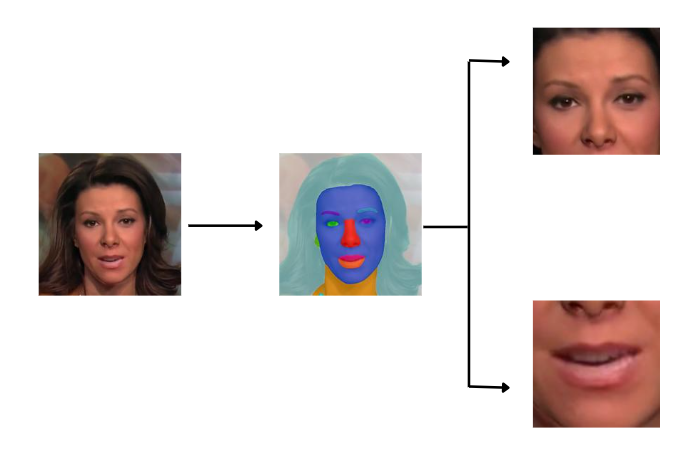} 
    \captionsetup{font=large} 
    \caption{Face Cropping.}
    \label{fig:sample_image}
\end{figure}

\begin{table}[htbp]
\centering
\renewcommand{\arraystretch}{1.5}
\begin{tabular}{|l|c|c|}
\hline
\textbf{Model } & \textbf{Validation Loss} & \textbf{Validation Accuracy} \\
\hline
FaceCropSwin & 0.2932 & 0.9156 \\
\hline
\end{tabular}
\vspace{3pt}
\caption{Model Performance on Face Crops}
\label{tab:data_summary}
\end{table}

The next step in the pipeline involves training two Swin Transformer models: one on the eyes and eyebrows crop and the other on the lips and chin crop. The final output is computed as the arithmetic mean of the probabilities generated by the two models. In cases where the face segmentation model fails to segment the image (e.g., due to missing facial parts), the output probability is set to 0.5 by default, indicating an indeterminate classification. This does not imply real or fake but suggests using a different model for such images instead of the one trained on face crops. This approach has been incorporated as a strength in our final ensemble model, particularly for handling challenging images where the person appears sideways or in unconventional poses.

\subsection{Models}
The following pipelines were trained on the entire dataset to improve deepfake detection performance.

\subsubsection{Vision Transformer (ViT)}
The Vision Transformer (ViT) \cite{b20} is utilized for this task by using its patch-based representation and self-attention mechanisms to extract meaningful features.

\begin{table}[htbp]
\centering
\renewcommand{\arraystretch}{1.5}
\begin{tabular}{|l|c|c|}
\hline
\textbf{Model } & \textbf{Validation Loss} & \textbf{Validation Accuracy} \\
\hline
ViT & 0.3883 & 0.8945 \\
\hline
\end{tabular}
\vspace{3pt}
\caption{Model Performance of ViT}
\label{tab:data_summary}
\end{table}

\subsubsection{ViT + Traditional Classifiers}
The Vision Transformer (ViT) combined with traditional machine learning classifiers enhances the robustness of deepfake detection. ViT is used to extract high-level features. These features are then validated using traditional classifiers such as xDNN (explainable deep neural network) \cite{b21}, Naive Bayes, KNN, and SVM, which serve as secondary layers for classification.

\begin{table}[htbp]
\centering
\renewcommand{\arraystretch}{1.5}
\begin{tabular}{|l|c|c|}
\hline
\textbf{Model } & \textbf{Validation Accuracy} \\
\hline
xDNN Validation & 0.9063 \\
\hline
\end{tabular}
\vspace{3pt}
\caption{Model Performance of xDNN}
\label{tab:data_summary}
\end{table}

\subsubsection{Xception}
Xception is a deep learning model based on depthwise separable convolutions, which replace standard convolution layers to improve computational efficiency and accuracy. It consists of entry, middle, and exit modules that process images hierarchically, capturing both spatial and channel-wise information. This makes it well-suited for deepfake detection tasks.

\begin{table}[htbp]
\centering
\renewcommand{\arraystretch}{1.5}
\begin{tabular}{|l|c|c|}
\hline
\textbf{Model } & \textbf{Validation Loss} & \textbf{Validation Accuracy} \\
\hline
Xception & 1.0552 & 0.7614 \\
\hline
\end{tabular}
\vspace{3pt}
\caption{Model Performance of Xception}
\label{tab:data_summary}
\end{table}

\subsubsection{EfficientNet-B7 with Squeeze-and-Expansion (SE) Blocks}
EfficientNet-B7, combined with Squeeze-and-Expansion (SE) blocks, provides a powerful framework for deepfake detection. EfficientNet-B7 employs compound scaling to balance network depth, width, and resolution, achieving high accuracy. The SE blocks enhance the model by recalibrating channel-wise feature responses, enabling it to focus on the most relevant features. This combination allows the model to capture intricate details and subtle artifacts in images, making it highly effective for deepfake detection.

\begin{table}[htbp]
\centering
\renewcommand{\arraystretch}{1.5}
\begin{tabular}{|l|c|c|}
\hline
\textbf{Model } & \textbf{Validation Loss} & \textbf{Validation Accuracy} \\
\hline
Xception & 0.6135 & 0.8600 \\
\hline
\end{tabular}
\vspace{3pt}
\caption{Model Performance of EfficientNet-B7 with SE blocks.}
\label{tab:data_summary}
\end{table}

\subsubsection{SimCLR}
SimCLR (Simple Framework for Contrastive Learning of Visual Representations) \cite{b17, b18} is a self-supervised learning framework adapted for deepfake detection. It consists of:
\begin{itemize}
    \item A base encoder, typically a convolutional neural network like ResNet \cite{b19}, which extracts feature representations from input images.
    \item A projection head, implemented as a multi-layer perceptron (MLP), maps the features into a lower-dimensional space suitable for contrastive learning.
    \item The model uses contrastive loss to make augmented versions of the same image more similar while pushing apart different images, enabling it to learn strong and consistent features. 
\end{itemize}

\begin{table}[htbp]
\centering
\renewcommand{\arraystretch}{1.5}
\begin{tabular}{|l|c|c|}
\hline
\textbf{Model } & \textbf{Validation Loss} & \textbf{Validation Accuracy} \\
\hline
SimCLR & 0.3883 & 0.8708 \\
\hline
\end{tabular}
\vspace{3pt}
\caption{Model Performance of SimCLR}
\label{tab:data_summary}
\end{table}

\subsubsection{\textbf{SFnet}}
The SFnet framework classifies images as \textit{``real"} or \textit{``fake"} by utilizing spatial and frequency domain features. It comprises three components: a Spatial Feature Extractor, a Frequency Domain Feature Extractor, and a Classification Head.

\begin{itemize}
    \item The Spatial Feature Extractor uses a Swin Transformer, generating a high-dimensional feature vector representing the image's structural and contextual information.
    \item The Frequency Domain Feature Extractor applies a 2D Fast Fourier Transform (FFT) to extract frequency-specific information. The magnitude (frequency intensity) and phase (spatial relationships) are processed through a CNN to produce a compact frequency feature vector.
    \item Both feature sets are concatenated and passed to the Classification Head, which consists of fully connected layers with non-linear activations and dropout. A sigmoid function outputs a probability score indicating whether the image is real or fake.
\end{itemize}

The model is trained using Binary Cross-Entropy Loss and optimized with the Adam optimizer. By combining spatial and frequency domain features, SFnet robustly detects subtle artifacts in fake images.

\subsubsection{\textbf{SFPnet}}
The SFPnet model is based on a Swin Transformer and extracts both spatial and frequency domain features from image patches.
Detecting deepfakes requires analyzing both pixel-level details and higher-order inconsistencies. Spatial features capture patterns like edges, textures, and object shapes, and expose visible manipulations. Frequency features reveal intensity and periodic artifacts. These highlight subtle inconsistencies such as compression or unnatural transitions. Using both spatial and frequency features improves cross-dataset detection. It does so by identifying shared frequency-domain artifacts across visually distinct forgeries and enhancing resilience to adversarial attacks, as modifying frequency artifacts without degrading quality is challenging.

\begin{itemize}
    \item The Swin Transformer outputs a feature vector for each patch, stored in the shape \([B, \text{num\_patches}, \text{spatial\_dim}]\), where \(B\) is the batch size, \(\text{num\_patches}\) is the number of patches, and \(\text{spatial\_dim}\) is the dimensionality of the spatial features.
    \item Frequency domain features are extracted by applying the Fast Fourier Transform (FFT) to each patch. The magnitude and phase of the frequency components are computed and passed through a small CNN-based encoder, producing a frequency feature vector of size \([B, \text{num\_patches}, \text{freq\_dim}]\).
    \item The spatial and frequency features are concatenated, resulting in a tensor of shape \([B, \text{num\_patches}, \text{spatial\_dim} + \text{freq\_dim}]\). An MLP-based aggregation method reduces the dimensionality and extracts a global image representation.
    \item The aggregated feature vector is passed through a Classification Head, which consists of an MLP with one hidden layer and GELU activation, producing the final classification output.
\end{itemize}

\subsubsection{\textbf{SwinAtten}}
The SwinAtten Attention model integrates four components: a Spatial Feature Extractor, a Frequency Domain Feature Extractor, a Patch Attention Mechanism, and a Classification Head.

\begin{itemize}
    \item The Spatial Feature Extractor uses a pre-trained Swin Transformer to extract high-level spatial features from the input image. It processes images in patches and outputs a feature vector of size \([B, \text{num\_patches}, \text{spatial\_dim}]\), where \(B\) is the batch size, \(\text{num\_patches}\) is the number of patches, and \(\text{spatial\_dim}\) is the dimensionality of the spatial features.
    \item The Frequency Domain Feature Extractor applies a 2D FFT to each patch, transforming the input image into the frequency domain. The magnitude and phase components are extracted and processed through a CNN, generating a frequency feature vector of size \([B, \text{num\_patches}, \text{freq\_dim}]\), where \(\text{freq\_dim}\) is the dimensionality of the frequency features.
    \item The Patch Attention Mechanism \cite{b16} combines the spatial and frequency feature vectors using a multi-head self-attention mechanism, producing a single feature vector of size \([B, \text{spatial\_dim} + \text{freq\_dim}]\).
    \item The Classification Head processes the aggregated feature vector through a multi-layer perceptron (MLP) with GELU activation, outputting a probability score between 0 and 1.
\end{itemize}

\begin{table}[htbp]
\centering
\renewcommand{\arraystretch}{1.5}
\begin{tabular}{|l|c|c|}
\hline
\textbf{Model } & \textbf{Validation Loss} & \textbf{Validation Accuracy} \\
\hline
Swin [Large] & 0.3122 & 0.9160 \\
\hline
SFnet & 0.2090 & 0.9199 \\
\hline
SFPnet & 0.2378 & 0.9235 \\
\hline
SwinAtten & 0.2313 & 0.9307 \\
\hline
\end{tabular}
\vspace{3pt}
\caption{Model Performances of various Swin Based models}
\label{tab:data_summary}
\end{table}

\section{Experiments and Results}
\begin{figure}[htbp]
    \centering
    \includegraphics[width=0.45\textwidth]{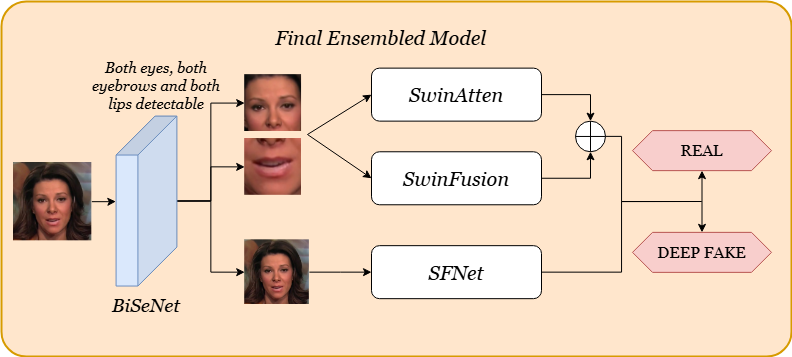} 
    \captionsetup{font=large} 
    \caption{The Final Pipeline.}
    \label{fig:sample_image}
\end{figure}
\subsection{Experiment Setup}
The experiments were conducted on an NVIDIA A100 GPU with 40 GB of memory, to handle the large compute and memory requirements of the models. The code is in Python, and the model architectures were constructed using PyTorch. To import pre-existing architectures (especially transformers), we used the TIMM library, which contains a wide variety of pretrained models, including the Swin transformer and its variants.

The experiments were conducted on the dataset provided by the competition organizers. Some experiments were conducted after processing and dividing the dataset to mitigate data imbalance issues as mentioned in the previous section. The images, being of 256x256 size, were resized (depending on the model), and normalized based on ImageNet statistics. We first generated benchmarks on the dataset using state-of-the-art models including ViT and Xception. Benchmarking gave a direction of progress and scope for improvement by focusing on certain characteristics of the dataset, images, and models. New architectures and pipelines were built by splitting the dataset, cropping the images, or combining different models and concepts (such as FFT and attention). Some models, such as SimCLR and SFnet, utilized data augmentation, while others did not. These individual pipelines and workflows were then evaluated on the validation set, in order to assess their suitability. Validation metrics used were validation loss, validation accuracy, F1, AUC, precision, recall, equal error rate (EER) and detection cost function (DCF).

Building new pipelines based on existing models did give an improvement in scores, however, they seemed to hit a ceiling and rarely crossed the 93-94\% validation accuracy mark. To try and improve the score further, we applied ensembling, i.e., combining the outputs of two or more models to give a final output. The idea is that one model's strengths will compensate for the other model's weaknesses, and multiple weak learners can together give a more robust model that performs well. Of all the pipelines tested, we chose 7 as candidates for ensembling (Table XI).

\begin{table*}[t]
\centering
\caption{Performance Metrics for Various Models Chosen for Ensembling}
\begin{tabular}{lccccccc}
\toprule
\textbf{Models} & \textbf{AUC} & \textbf{Accuracy} & \textbf{F1-Score} & \textbf{Precision} & \textbf{Recall} & \textbf{EER} & \textbf{DCF} \\
\midrule
FaceCropSwin     & 0.9353 & 0.8519 & 0.8393 & 0.9090 & 0.7795 & 0.0969 & 0.1558 \\
EfficientNet-B7  & 0.9611 & 0.8708 & 0.8802 & 0.8146 & \textbf{0.9573} & 0.1027 & 0.1140 \\
SFnet            & \textbf{0.9807} & \textbf{0.9349} & \textbf{0.9344} & \textbf{0.9338} & 0.9350 & \textbf{0.0659} & \textbf{0.0656} \\
SimCLR           & 0.9683 & 0.8711 & 0.8805 & 0.8151 & \textbf{0.9573} & 0.0743 & 0.1138 \\
ViT              & 0.9734 & 0.9020 & 0.9062 & 0.8629 & 0.9541 & 0.0782 & 0.0915 \\
SwinAtten        & \textbf{0.9799} & \textbf{0.9372} & \textbf{0.9375} & \textbf{0.9258} & 0.9495 & \textbf{0.0627} & \textbf{0.0624} \\
SwinFusion       & \textbf{0.9814} & \textbf{0.9404} & \textbf{0.9411} & \textbf{0.9236} & \textbf{0.9593} & \textbf{0.0478} & \textbf{0.0586} \\
\bottomrule
\end{tabular}
\label{tab:performance_metrics}
\end{table*}

Note, the tables in the previous section use a threshold of 0.5. However, upon closer inspection, we noticed that using a threshold of 0.5, the models perform well on fake images, while comparatively underperforming on real images in general. Assuming the convention of real scores being higher, this suggested that the models were underconfident on real images, and to mitigate this, we changed the threshold, choosing a value of 0.3, after testing various values. 

In other words, the table above uses a threshold of 0.3. This means that scores below 0.3 indicate "fake," while scores above 0.3 indicate "real." This adjustment ensures the model predicts "fake" only when it is very confident, and "real" otherwise. The primary reason for this adjustment is the data imbalance, where an inherent bias towards "fake" skews the model's perception of images. By setting the threshold to 0.3, we essentially require stronger evidence to classify an image as fake.

The results of the experiments, as shown in Table \ref{tab:performance_metrics}, highlight the performance of various models chosen for ensembling. Among the models, SwinFusion achieved the highest accuracy (94.04\%) and AUC (98.14\%), demonstrating its effectiveness in detecting deepfakes. SwinAtten and SFnet also performed well, with accuracies of 93.72\% and 93.49\%, respectively. These models used transformer-based architectures and frequency-domain features, which proved effective in capturing subtle artifacts in manipulated images. On the other hand, FaceCropSwin and EfficientNet-B7 showed relatively lower performance, with accuracies of 85.19\% and 87.08\%, respectively, indicating that simpler architectures or those without specialized feature extraction mechanisms struggled with the complexity of the dataset.

Upon further experimenting with ensembling types, including majority voting, averaging, and weighted average, the final pipeline (refer Fig \ref{fig:sample_image}) took shape, comprising of a BiSeNet, SFNet, SwinAtten and SwinFusion, combining the best performing models, improving performance and robustness while reducing the number of training parameters. The pipeline is as follows:
\begin{itemize}
    \item The image passes through a pretrained BiSeNet (as it does for the FaceCropSwin pipeline). The BiSeNet detects whether both the eyebrows, both the eyes and both the lips are detectable in the image. The difference is that this time, no cropping of the image takes place.
    \item If all 6 components are detectable, the input image is passed to a SwinAtten and a SwinFusion model in parallel. Their scores are averaged before applying the threshold.
    \item If at least one of the components cannot be detected, the input image is passed to an SFnet model.
\end{itemize}

The metrics from the pipeline are listed in Table \ref{tab:single_column_metrics}. Processing one image on our pipeline takes 0.417 seconds and close to 425M trainable and non-trainable parameters. 

\begin{table}[t]
\centering
\caption{Performance Metrics on Final Architecture}
\begin{tabular}{lc}
\toprule
\textbf{Metric}     & \textbf{Value} \\
\midrule
AUC                 & \textbf{0.9822}         \\
Accuracy            & \textbf{0.9613}         \\
F1-Score            & \textbf{0.9609}         \\
Precision           & \textbf{0.9619}         \\
Recall              & \textbf{0.9600}         \\
EER                 & \textbf{0.0388}         \\
DCF                 & \textbf{0.0391}         \\
\bottomrule
\end{tabular}
\label{tab:single_column_metrics}
\end{table}

\section{Conclusion}
We introduced a novel ensemble framework for deepfake detection that combines transformer-based architectures such as Swin Transformers and Vision Transformers (ViTs) with texture-based methods. The proposed approach has innovative techniques like data-splitting based on human features, sequential training, frequency splitting, patch-based attention, and face segmentation to overcome issues like dataset imbalance and subtle artifacts in critical regions such as the eyes and mouth. State-of-the-art performance was achieved by the ensemble model on the DFWild-Cup dataset, where it showed good accuracy, robustness, and generalization capability over a variety of deepfake generation techniques. Our framework can be considered to be a very reliable solution in the real world, addressing the problem of deepfakes that become increasingly realistic in synthetic media.

Future work includes improving the dataset by adding more types of deepfakes and balancing real and fake samples. We will also work on the speed of the model for real-time use and test it on various datasets to make sure it performs well in all scenarios. Video analysis and simplification of the model will make it more accurate and trustworthy. Finally, we will analyze how the model performs when attacked to try to fool it, thus making it more reliable in real-world applications. These steps will help create better tools to detect and stop fake media.

\end{document}